\documentclass{article}

\PassOptionsToPackage{sort&compress}{natbib}

\usepackage{xparse}
\usepackage{soulutf8}
\usepackage[colorinlistoftodos]{todonotes}
\DeclareRobustCommand{\hlc}[2]{{\sethlcolor{#1}\hl{#2}}}
\makeatletter
\NewDocumentCommand{\NewCommentator}{mm}
{%
  \expandafter\NewDocumentCommand\csname\expandafter\@gobble\string #1\endcsname{smm}{%
   \IfBooleanTF{##1}{\hlc{#2}{##2}\todo[color=#2,size=\tiny]{##3}}{\textcolor{#2}{\st{##2}}~\textcolor{#2}{##3}}%
   }%
}
\makeatother

\usepackage[preprint]{corl_2021} 

\usepackage{url}
\usepackage{graphicx}
\usepackage{amsmath}
\usepackage{amsfonts}
\usepackage{mathtools}  
\usepackage{float}
\usepackage{siunitx}
\usepackage{subcaption}
\usepackage{arydshln}
\usepackage{wrapfig}
\usepackage[normalem]{ulem}

\newcommand{\norm}[1]{\left\lVert#1\right\rVert}

\newcommand{\lhat}[0]{\mathbf{l}}
\newcommand{\m}[0]{\mathbf{m}}
\newcommand{\bm}[1]{\mathbf{#1}}

\NewCommentator{\rl}{orange}

\DeclareMathOperator{\Tr}{tr}
\DeclareMathOperator{\diag}{diag}

\title{Distributional Depth-Based Estimation \\ of Object Articulation Models}

\author{
  Ajinkya Jain\thanks{Corresponding author: \texttt{ajinkya@utexas.edu}}\\
  UT Austin \\
  \And
  Stephen Giguere\thanks{Equal contribution, presented alphabetically}\\
  UT Austin \\
  \And
  Rudolf Lioutikov\footnotemark[2] \\
  Karlsruhe Institute of Technology \\
  \And
  Scott Niekum\\
  UT Austin \\
}

\begin{document}
\maketitle


\begin{abstract}
We propose a method that efficiently learns distributions over articulation model parameters directly from depth images without the need to know articulation model categories a priori. By contrast, existing methods that learn articulation models from raw observations typically only predict point estimates of the model parameters, which are insufficient to guarantee the safe manipulation of articulated objects. Our core contributions include a novel representation for distributions over rigid body transformations and articulation model parameters based on screw theory, von Mises-Fisher distributions, and Stiefel manifolds. Combining these concepts allows for an efficient, mathematically sound representation that implicitly satisfies the constraints that rigid body transformations and articulations must adhere to. Leveraging this representation, we introduce a novel deep learning based approach, DUST-net, that performs category-independent articulation model estimation while also providing model uncertainties. We evaluate our approach on several benchmarking datasets and real-world objects and compare its performance with two current state-of-the-art methods. Our results demonstrate that DUST-net can successfully learn distributions over articulation models for novel objects across articulation model categories, which generate point estimates with better accuracy than state-of-the-art methods and effectively capture the uncertainty over predicted model parameters due to noisy inputs. \href{https://pearl-utexas.github.io/DUST-net/}{[webpage]}

\end{abstract}



\section{Introduction}
Articulated objects, such as drawers, staplers, refrigerators, and dishwashers, are ubiquitous in human environments. These objects consist of multiple rigid bodies connected via mechanical joints such as hinge joints or slider joints. Robots in human environments will need to interact with these objects often while assisting humans in performing day-to-day tasks. To interact safely with such objects, a robot must reason about their articulation properties while manipulating them. 
%
An ideal method for learning such properties might estimate these parameters directly from raw observations, such as RGB-D images while requiring limited or no a priori information about the task. The ability to additionally provide a confidence over the estimated properties, would allow such a method to be leveraged in the development of safe motion policies for articulated objects~\citep{jain2018efficient}.

The majority of existing methods to learn articulation models for objects from visual data either need fiducial markers to track motion between object parts~\citep{sturm2011probabilistic, katz2008manipulating, katz2013interactive, niekum2015online} or require textured objects~\citep{pillai2015learning, martin2014online, martin2016integrated, martin2019coupled, jain2019learning}. 
%
%
Recent deep-learning based methods address this by predicting articulation properties for objects from raw observations, such as depth images~\citep{abbatematteo2019learning, li2020category,  liu2020nothing, jain2020screwnet} or PointCloud data~\citep{wang2019shape2motion, yan2019RPM}. However, the majority of these methods~\citep{abbatematteo2019learning, li2020category, wang2019shape2motion, yan2019RPM} require knowledge of the articulation model category for the object (e.g., whether it has a revolute or prismatic joint) which may not be available in many realistic settings.
Alleviating this requirement,~\citet{jain2020screwnet} introduced ScrewNet, which uses a unified representation based on screw transformations to represent different articulation types and performs category-independent articulation model estimation directly from raw depth images. 
However, ScrewNet~\citep{jain2020screwnet} and related methods~\citep{abbatematteo2019learning, li2020category, liu2020nothing, wang2019shape2motion, yan2019RPM} only predict point estimates for an object's articulation model parameters. Nonetheless, reasoning about the uncertainty in the estimated parameters can provide significant advantages for ensuring success in robot manipulation tasks, and allows for further advancements such as robust planning~\citep{jain2018efficient}, active learning using human queries~\citep{cui2018active}, and the learning of behavior policies that provide safety assurances~\citep{taylor20safe}. Motivated by these advantages, we propose a method for learning articulation models, which estimates the uncertainty over model parameters using a novel distribution over the set of screw transformations based on the matrix von Mises-Fisher distribution over Stiefel manifolds~\citep{chikuse2003statistics}.
We introduce DUST-net, \textbf{D}eep \textbf{U}ncertainty estimation on \textbf{S}crew \textbf{T}ransforms-\textbf{net}work, a novel deep learning-based method that, in addition to providing point estimates of the object's articulation model parameters, leverages raw depth images to provide uncertainty estimates that can be used to guide the robot's behavior without requiring to knowledge of the object's articulation model category a priori.


DUST-net garners numerous benefits over existing methods. First, DUST-net estimates articulation properties for objects with uncertainty estimates, unlike most current methods~\citep{abbatematteo2019learning, li2020category, jain2020screwnet, liu2020nothing, wang2019shape2motion, yan2019RPM}. These uncertainty estimates, apart from helping robots to manipulate objects safely~\citep{jain2018efficient}, could allow a robot to take information-gathering actions when it is not confident and enhance its chances of success in completing the task. Second, 
similar to ScrewNet~\citep{jain2020screwnet},
DUST-net can estimate model parameters without the need to to know the articulation model category a priori, by leveraging the unified representation for different articulation model types. Third, this unified representation helps DUST-net to be more computationally and data-efficient than other state-of-the-art methods~\citep{abbatematteo2019learning, li2020category}, as it uses a single network to estimate model parameters for all common articulation models, unlike other methods that require a separate network for each articulation model category~\citep{abbatematteo2019learning, li2020category, wang2019shape2motion, yan2019RPM}. Empirically, DUST-net outperforms other methods even when trained using only half the training data in comparison. Fourth, the distributional learning setting yields more robustness to outliers and noise. Fifth, DUST-net is able to reliably estimate distributions over articulation model parameters for objects in the robot's camera frame. By contrast, ScrewNet~\citep{jain2020screwnet}, the most closely related approach to ours, can only predict point estimates for articulation model parameters in the object's local frame. 



\begin{figure}[t]
\centering
    \includegraphics[width=0.85\linewidth, height=0.45\linewidth]{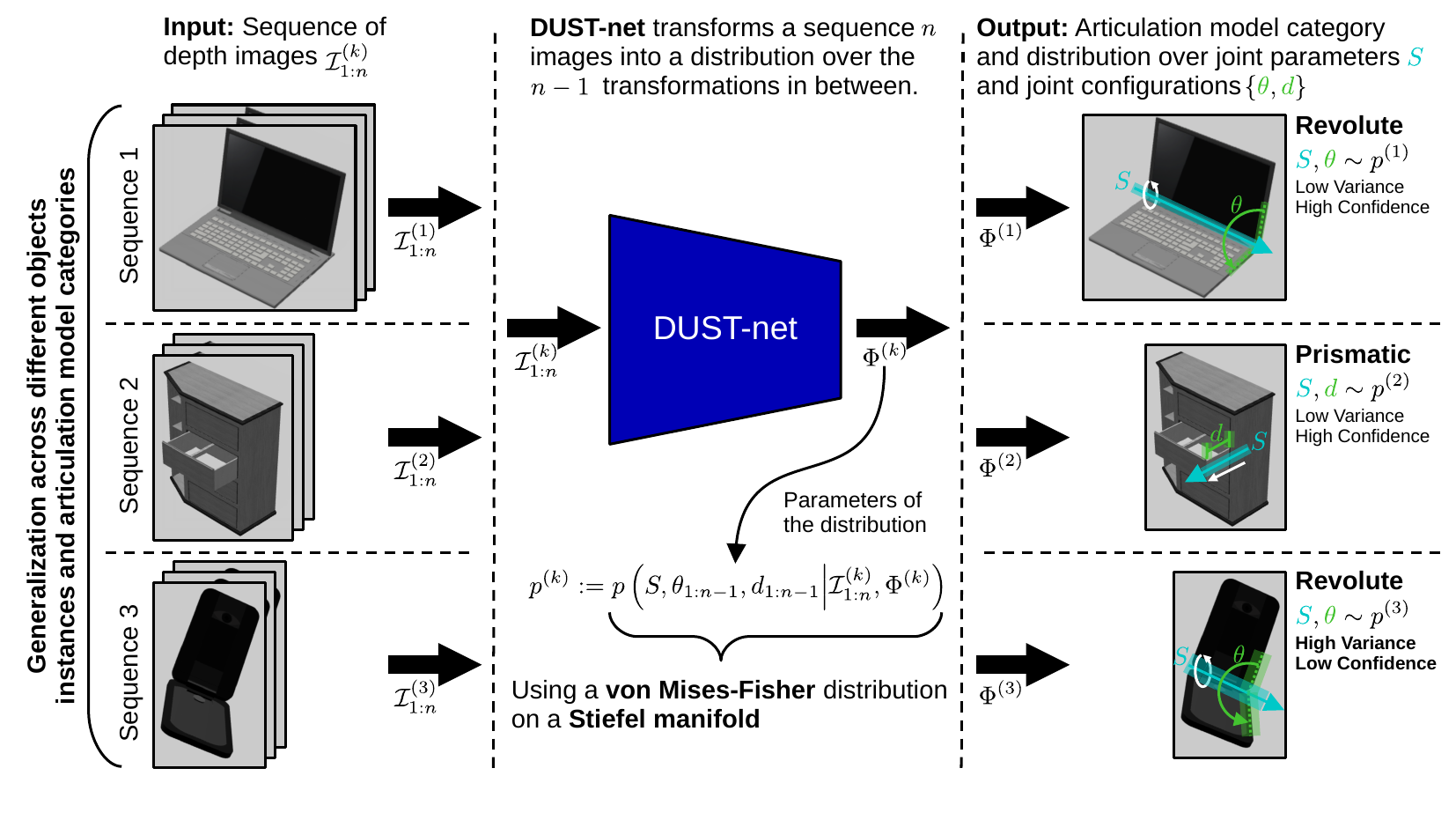}
    \caption{DUST-net uses a sequence of images $\mathcal{I}_{1:n}$ to compute the parameters, $\Phi$, of the conditional distribution over the joint parameters $S$ and configurations $\{\theta,d\}_{1:n-1}$. This distribution allows for inference and reasoning, such as uncertainty and confidence, over both the parameters and the configurations. Using a von Mises-Fisher distribution on a Stiefel manifold allows for an efficient reparameterization that inherently obeys multiple constraints that define rigid body transformations.}
    \label{fig:overview}
    \vspace{-5pt}
\end{figure}

We evaluate DUST-net through experiments on two benchmarking datasets: a simulated articulated objects dataset \cite{abbatematteo2019learning} and the PartNet-Mobility dataset \cite{Xiang_2020_SAPIEN, Mo_2019_CVPR, chang2015shapenet}, as well as three real-world objects: a microwave, a drawer, and a toaster oven. We compare DUST-net with two state-of-the-art methods, namely ScrewNet~\citep{jain2020screwnet} and an MDN-based method proposed by~\citet{abbatematteo2019learning}, as well as two baseline methods. The experiments demonstrate that the samples drawn from the distributions learned by DUST-net result in significantly better estimates for articulation model parameters in comparison to the point estimates predicted by other methods. Additionally, the experiments show that DUST-net can successfully and accurately capture the uncertainty over articulation model parameters resulting from noisy inputs.



\section{Related Work}
\label{sec:litsurvey}
\vspace{-2pt}

\textbf{Articulation model estimation from visual observations:} A widely used approach for estimating articulation models is based on the probabilistic framework proposed by~\citet{sturm2011probabilistic}. It uses the time-series observations of 6D poses of different parts of an articulated object to learn the relationship between them~\citep{sturm2011probabilistic, niekum2015online, pillai2015learning, jain2019learning}.
More recently,~\citet{abbatematteo2019learning} and \citet{li2020category} proposed methods to learn articulation properties for objects from raw depth images given articulation model category. In a related body of work on object parts mobility estimation,~\citet{wang2019shape2motion} and~\citet{yan2019RPM} proposed approaches to segment different parts of the object in an input point cloud and estimate their mobility relationships, given a known articulation model category. Alleviating the requirement of having a known articulation model category, ~\citet{jain2020screwnet} recently proposed ScrewNet that performs category-independent articulation model estimation from depth images.
However, these methods only predict point estimates for the articulation model parameters, while DUST-net predicts a distribution over their values.

\textbf{Rigid Body Pose Estimation}: 
Our contributions are related to existing work on estimating distributions describing the orientation of rigid bodies.
~\citet{gilitschenski2015unscented}, ~\citet{arun2018probabilistic},  ~\citet{srivatsan2016estimating} and \citet{rosen2019se} propose strategies that can be used to estimate the rigid body transformation of an object using a combination of Bingham and Gaussian distributions, and the von Mises-Fisher distribution, respectively.
The mathematical model used by our approach is inspired by these works, but 1) extends them to also represent uncertainty over the configuration of articulated object components about screw axes, and 2) integrates them into a deep learning model that is capable of learning these configurations from raw depth images.
In addition, while these approaches use distributions over orientations and rigid body transformations to produce estimates, DUST-net directly outputs a distribution that can be used to facilitate further applications such as uncertainty-aware behavior planning.
%

\textbf{Interactive perception (IP)}:~\citet{katz2008manipulating} introduced IP as a method to leverage a robot's interaction with objects to generate a rich perceptual signal for articulation model estimation for planar objects, and extended it to learn 3D articulation models for objects~\citep{katz2013interactive}.~\citet{martin2016integrated} used hierarchical recursive Bayesian filters to make estimation more robust and developed online methods for articulation model estimation from RGB images~\citep{martin2014online, martin2016integrated, martin2019coupled}. A comprehensive survey on IP methods in robotics was presented by~\citet{bohg2017interactive}. While IP presents a powerful tool for estimating articulation properties for objects, a wide majority of existing IP methods require textured objects, unlike DUST-net, which learns these properties using depth images. 


\textbf{Further approaches}:
Articulation motion models can be viewed as geometric constraints imposed on multiple rigid bodies. Such constraints can be learned from human demonstrations by leveraging different sensing modalities~\citep{perez2017c, liu2019learning, daniele2020multiview, liu2020nothing, subramani2018inferring}. Recently,~\citet{daniele2020multiview} proposed a multimodal learning framework that incorporates both vision and natural language information for articulation model estimation. However, these approaches predict point estimates for the articulation model parameters, unlike DUST-net, which predicts a distribution over the articulation model parameters.


\section{Problem Formulation:} 
Given a sequence of $n$ depth images $\mathcal{I}_{1:n}$ of motion between two parts of an articulated object, we estimate the parameters of a probability distribution $p(\phi | \mathcal{I}_{1:n})$ representing uncertainty over the parameters $\phi$ of the articulation model $\mathcal{M}$ governing the motion between the two parts. Following \citet{jain2020screwnet}, we define the model parameters $\phi$ as the parameters of the screw axis of motion, $\mathsf{S} = (\lhat, \m)$, where both $\lhat$ and $\m$ are elements of $\mathbb{R}^3$. This unified parameterization can be used in articulation models with at most one degree-of-freedom (DoF), namely rigid, revolute, prismatic, and helical~\citep{jain2020screwnet}. Additionally, we estimate the parameters of a distribution $p(q_{1:n-1} | \mathcal{I}_{1:n})$ representing uncertainty over the configurations $q_{1:n-1}$ identifying the rigid body transformations between the two parts in the given sequence of images $\mathcal{I}_{1:n}$ under model $\mathcal{M}$ with parameters $\phi$. Configurations $q_i, i \in \{1...n-1\}$ correspond to a set of tuples, $q_i = (\theta_i, d_i)$, defining a rotation around and a displacement along the screw axis $\mathsf{S}$\footnote{Please refer to the supplementary material for further details}. 
We assume that the relative motion between the two object parts is determined by a single articulation model.  

\vspace{-3pt}
\section{Approach}
\vspace{-3pt}
\label{sec:approach}
Given a sequence of depth images $\mathcal{I}_{1:n}$ of motion between two parts of an articulated object, DUST-net estimates parameters of the joint probability distribution $p(\phi, q_{1:n-1} | \mathcal{I}_{1:n})$ representing uncertainty over the articulation model parameters $\phi$ governing the motion between the two parts and the observed configurations $q_{1:n-1}$. 
When deciding how to learn this distribution, two goals arise.
While some parameters, such as the translation of an object part along a screw axis, are defined on Euclidean space, the set of valid screw axes exhibits constraints that prevent standard distributions defined on $\mathbb{R}^6$ from being applied without complicating the learning process.
For example, a standard representation for distributions over screw axes can be the product of a Bingham distribution over the line's orientation and a multivariate normal distribution over its position in space~\citep{siciliano2016springer}. However, this representation produces non-unique estimation targets. A rotation of $\theta$ about the screw axis with orientation $\lhat$ results in the same transformation as a rotation of $-\theta$ about the screw axis with orientation $-\lhat$. Similarly, a displacement $d$ along $\lhat$ results in the same transformation as a displacement $-d$ along $-\lhat$. This leads to ambiguities in the targets in the estimation problem and can hinder the performance of the trained estimator.
By selecting a representation that accounts for these symmetries, these non-unique estimation targets are removed.
Second, once a suitable parameterization is chosen, we seek a parametric form for the joint distribution which can be learned by a deep network.
%

First, we consider the problem of parameterizing the set of screw axes.
As noted earlier, we define the model parameter $\phi$ as the parameters of the screw axis of motion $\mathsf{S} = (\lhat, \m)$. However, this parameterization requires that $\lhat$ has unit norm, and that $\lhat$ and $\m$ are orthogonal. To eliminate these constraints, we rewrite the moment vector of a screw axis as $\m = \norm{\m} \hat{\mathbf{m}}$, where $\norm{\m}$ and $\hat{\mathbf{m}}$ represent its magnitude and a unit vector along it respectively, and the Pl\"{u}cker coordinates for the screw axis as $\mathsf{S} = (\lhat, \hat{\mathbf{m}}, \norm{\m})$. The Pl\"{u}cker coordinates can then be seen as an unconstrained point in the space $\mathbb{S} := \mathrm{V}_{2, 3} \times \mathbb{R}^+$, where $(\lhat, \hat{\mathbf{m}}) \in \mathrm{V}_{2, 3}$ with $\mathrm{V}_{2, 3}$ denoting the \emph{Stiefel manifold} of 2-frames in $\mathbb{R}^3$ 
and $\norm{\m} \in \mathbb{R}^+$ with $\mathbb{R}^+$ denoting the set of positive real numbers. 
The Stiefel manifold $\mathrm{V}_{k, m}$ is the space whose points are sets of $k$ orthonormal vectors in $\mathbb{R}^m$, called $k$-frames in $\mathbb{R}^m~(k \leq m)$\footnotemark[1] \footnotetext[1]{Please refer to the supplementary material for further details}~\citep{chikuse2003statistics}.
Consequently, because of the one-to-one mapping from elements of $\mathrm{V}_{2, 3} \times \mathbb{R}^+$ to screw axes, the non-unique estimation targets described above are eliminated. 
Based on this parametrization of screw axes, we define the set of valid configuration parameters as follows.  We restrict the range of values for the rotation about the screw axis to be $\theta \in [0, 2\pi)$ and restrict the displacement along the axis to be $d \in \mathbb{R}^+$. Note that these constraints do not reduce the representational power of the screw transform $(\lhat, \m , \theta, d)$ to denote a general rigid body transform, but merely ensure a unique representation.


Having described the parameterization of the set of screw axes and configurations, we now consider the task of defining a joint probability distribution over their values.
We propose to represent the distribution over predicted screw axis parameters, $p(\mathsf{S} ~|~ \mathcal{I}_{1:n})$ with $\mathsf{S} \in \mathbb{S}$, as a product of a matrix von Mises-Fisher distribution $\mathcal{F}(\cdot | 3, \mathbf{F})$ defined on the Stiefel manifold $\mathrm{V}_{2, 3}$\footnotemark[1] and a truncated normal distribution $\mathcal{N}^+(\cdot | \mu, \sigma)$ with truncation interval $[0, +\infty)$ over $\mathbb{R}^+$. 
Formally,
\begin{equation}
    p(\mathsf{S} ~|~ \mathcal{I}_{1:n}) = p\left(\,l,\hat{\m},\norm{\m} ~\middle|~ \mathcal{I}_{1:n}, \bm{F}, \mu_\m, \sigma^2_\m\right) = \mathcal{F}\left(\,\lhat, \hat{\mathbf{m}} ~\middle|~ 3,\bm{F}\right)~\mathcal{N}^+\left(\,\norm{\m}~|~ \mu_\m, \sigma^2_\m\right),
    \label{eq:screw_dist}
\end{equation}
where \textbf{F} is a $3 \times 2$ matrix representing the parameters of the matrix von Mises-Fisher distribution over $\mathrm{V}_{2, 3}$, and $\mu_\m$ and $\sigma_\m$ denote the mean and standard deviation of the truncated normal distribution. 

Given the sequence of $n$ images, we also wish to estimate the posterior over configurations
$q_{1:n-1} = \{ \theta_{1:n-1}, d_{1:n-1}\}$ corresponding to the rotations about and displacements along the screw axis $\mathsf{S}$. We define the joint posterior representing the uncertainty 
over the screw axis $\mathsf{S}$ and the configurations $\{ \theta_{1:n-1}, d_{1:n-1}\}$ about it as a product of the aforementioned distribution and a set of distributions defined over the configuration parameters,
\begin{equation}
    p(\mathsf{S}, \theta_{1:n-1}, d_{1:n-1} ~|~ \mathcal{I}_{1:n}, \Phi) =  p(\mathsf{S} ; \bm{F}, \mu_\m, \sigma^2_\m)~\Psi(\theta_{1:n-1}; \psi)~ \Upsilon(d_{1:n-1}; \upsilon)
    \label{eq:param_dist}
\end{equation}
where $\Phi = \{ \bm{F}, \mu_\m, \sigma^2_\m, \psi, \upsilon\}$ is the set of parameters for the distribution and $\Psi$ and $\Upsilon$ represent the set of distributions having parameters $\psi$ and $\upsilon$ over the configurations $\theta_{1:n-1}$ and $d_{1:n-1}$, respectively. 
For the sake of brevity, we present further details on modeling assumptions in the supplementary material (see Appendix~\ref{sec:app_model}).
In this work, we consider $\Psi$ and $\Upsilon$ to be products of truncated normal distributions such that $\Psi = \prod^{n-1}_{i=1} \mathcal{N}^+(\theta_i|\text{M}^i_\theta, \sigma^2_\theta)$ and $\Upsilon = \prod^{n-1}_{i=1} \mathcal{N}^+(d_i|\text{M}^i_d, \sigma^2_d)$ with $\text{M}_\theta = \{\mu^1_\theta, ..., \mu^{n-1}_\theta \}$, $\text{M}_d = \{\mu^1_d, ..., \mu^{n-1}_d \}$, $\sigma_\theta$, and $\sigma_d$ denoting the set of means and the standard deviations of the set of truncated normal distributions over the configurations $\theta_{1:n-1}$ and $d_{1:n-1}$, respectively.

\textbf{Distribution parameter matrix F:} The parameter matrix for the matrix von Mises-Fisher distribution over $\mathrm{V}_{3,2}$ is a $3 \times 2$ matrix, $\bm{F}$. This presents two possible parameterizations of the matrix: first, to estimate each of the 6 elements of the $3 \times 2$ matrix \textbf{F} and second, to estimate the matrices $\Gamma, \Lambda$, and $\Omega$ defining the SVD of $\bm{F}$, given by $\bm{F} = \Gamma \Lambda \Omega^T$. The second parameterization  decouples the two objectives of distribution mode alignment with the ground truth labels and uncertainty representation; the mode of the distribution is given by $M = \Gamma \Omega^T$, and the concentration matrix for the distribution is given by $K = \Omega \Lambda \Omega^T$. This decoupling allows the network to independently optimize both objectives, whereas in the first parameterization, changes in the elements of $\bm{F}$ causes changes in both components.

By definition, $\Lambda$ is a $2\times 2$ diagonal matrix with two independent parameters, and $\Omega \in O(2)$ is a rotation matrix in two dimensions with one independent parameter, the rotation angle $\omega$. The matrix $\Gamma \in \tilde{\mathrm{V}}_{3,2}$ can be constructed from a rotation matrix $\text{R} \in O(3)$ by keeping only the first two columns of R. Hence, the matrix $\Gamma$ can be defined by three independent Euler angles, $(\alpha, \beta, \gamma)$ denoting rotation according to the $ZYX$ convention in the rotating frame. Euler angles can suffer from the problem of gimble lock~\citep{siciliano2016springer}, which we resolve by restricting the Euler angles to be in the ranges $\alpha \in [0, 2\pi), \beta \in [0, \pi)$, and $\gamma \in [0, 2\pi)$. 

\textbf{Normalization factor:} One of the main challenges of using the matrix von Mises-Fisher distribution is the calculation of its normalization factor ${}_0F_{1} (\frac{m}{2}, \frac{1}{4} \Lambda^2)$, which is a hypergeometric function of matrix argument \citep{chikuse2003statistics}. In this work, we approximate this hypergeometric function using a truncated series in terms of zonal polynomials, which are multivariate symmetric homogeneous polynomials and form a basis of the space of symmetric polynomials~\citep{chikuse2003statistics}. Through our preliminary experiments, we found that this truncated series is a good approximation of ${}_0F_{1}$ as it converges to a finite value, if the singular values of the $F$, i.e. $\lambda_1$ and $\lambda_2$ 
 are less than $ \lambda_{max}=50$.

\textbf{Architecture:} DUST-net sequentially connects a ResNet-18 CNN \citep{he2016deep} and a 2-layer MLP. ResNet-18 extracts task-relevant features from the input images, which are used by the MLP to predict a set of parameters $\Phi$ for the distribution $p(\mathsf{S}, \theta_{1:n-1}, d_{1:n-1} ~|~ \mathcal{I}_{1:n}, \Phi)$. The network is trained end-to-end, with ReLU activations for the hidden fully-connected layers. The first four output (out of 40) of the last linear layer of MLP, corresponding to the parameters $(\alpha, \beta, \gamma)$ and $\omega$ representing the matrices $\Gamma$ and $\Omega$ respectively, are fed through a ReLU-6 layer to ensure that the predictions map to their respective ranges. Remaining output is fed through a Softplus layer for non-negative output. 
Detailed network architecture is presented in the appendix (Fig.~\ref{fig:arch}).


 

\textbf{Training:} The training data for the model consists of sequences of depth images of objects parts moving relative to each other and the corresponding screw transforms $\mathbf{y} = (l,\hat{\m},\norm{\m}, \theta_{1:n-1}, d_{1:n-1})$. The objects and depth images are rendered in Mujoco \citep{todorov2012mujoco}. We train DUST-net by maximizing the log-probability of the labels $\mathbf{y}$ under the distribution  $p(\mathbf{y} ~|~ \mathcal{I}_{1:n}, \Phi)$: ~$\mathcal{L}(\bm{y}, \Phi) = -\log p(\bm{y}~|~\Phi)$. 
We assume that the observed configurations in $\mathcal{I}_{1:n}$ share the same variance. We use the precision parameters rather than the standard deviations, $\sigma_\m, \sigma_\theta$ and $\sigma_d$ to represent the distribution during training for better numerical stability. Following the discussion on training MDNs by \citet{makansi2019overcoming}, we separate the training in three stages. In the first stage, we assume the dispersion of the matrix von Mises-Fisher distribution to be fixed with $\Lambda = \diag(\lambda_0, \lambda_0),~\lambda_0 = 1$ and learn parameters corresponding to $\Gamma$ and $\Omega$ matrices. In the second stage, we fix the $\Lambda$ matrix and learn the rest of the parameters in the set $\Phi$. Finally, we train to predict the complete set $\Phi$. 

\begin{figure}[b]
    \centering
    \includegraphics[width=0.85\linewidth, height=0.33\linewidth]{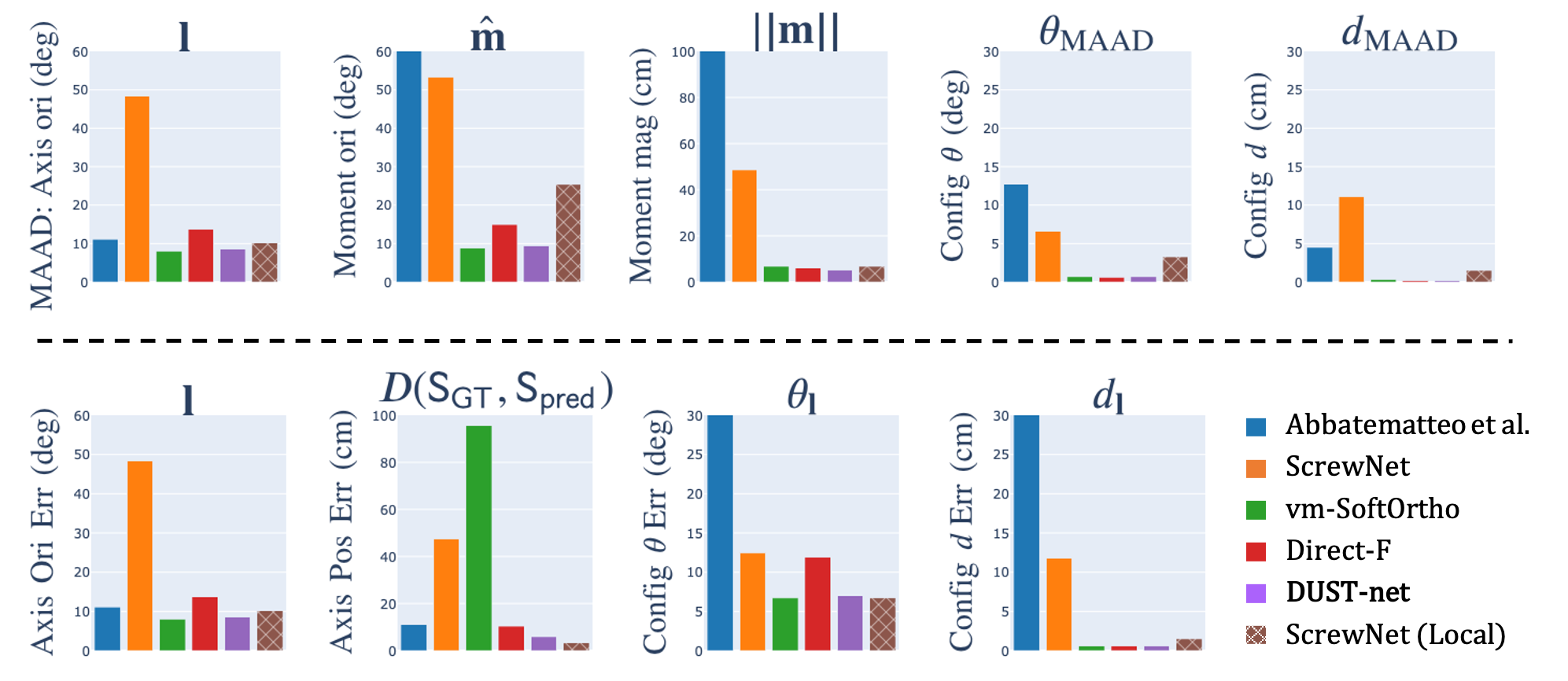}
    \caption{Mean error values on the MAAD (top) and Screw Loss (bottom) metrics for the simulated articulated objects dataset~\citep{abbatematteo2019learning} (lower values are better).
    Point estimates for DUST-net (violet) correspond to the modes of the distributions predicted by DUST-net.
    }
    \label{fig:pt_synart}
\end{figure}


\vspace{-2pt}
\section{Experiments}
\label{sec:results}
\vspace{-2pt}
In this section, we evaluate DUST-net on its ability to learn articulation model parameters and uncertainty estimates. 
We conducted three sets of experiments evaluating DUST-net's performance under different criteria: (1) how accurate point estimates of the articulation model parameters drawn from DUST-net's estimated distribution are in comparison to the existing methods, (2) how effectively DUST-net captures the uncertainty over parameters arising from noisy input, 
and (3) how effectively DUST-net transfers from simulation to a real-world setting.
We evaluated DUST-net's performance on two simulated benchmarking datasets: the objects dataset provided by Abbatematteo et al. \cite{abbatematteo2019learning}, and the PartNet-Mobility dataset \cite{Xiang_2020_SAPIEN, Mo_2019_CVPR, chang2015shapenet}, as well as a set of three real-world objects. 
From the simulated articulated object dataset \cite{abbatematteo2019learning}, we considered the cabinet, microwave, and toaster oven for revolute articulations and the drawer object class for prismatic articulations. 
From the PartNet-Mobility dataset \cite{Xiang_2020_SAPIEN, Mo_2019_CVPR, chang2015shapenet}, we considered five object classes: the dishwasher, oven, and microwave object classes for the revolute articulation model category, and the storage furniture object class consisting of either a single column of drawers or multiple columns of drawers, for the prismatic articulation model category. Among the three sets of experiments, we conducted the first two sets of experiments on the simulated datasets, while the last set of experiments were conducted on the real-world object dataset. In all the experiments, we assumed that the input depth images are semantically segmented and contain non-zero pixels corresponding only to the two objects between which we wish to estimate the articulation model.

We compared DUST-net's performance in estimating point estimates for articulation model parameters with two state-of-the-art methods, ScrewNet~\citep{jain2020screwnet} and an MDN-based approach proposed by~\citet{abbatematteo2019learning}. ScrewNet estimates the object's articulation model parameters in a local frame located at the center of the object, whereas DUST-net does so directly in the camera frame. We compare our method with ScrewNet predicting parameters both in the object local frame and the camera frame. Additionally, we propose two baseline methods that estimate distributions over articulation model parameters and compare to them DUST-net. The first baseline method (vm-SoftOrtho) can be viewed as an extension of ScrewNet to a distributional setting. It represents the uncertainty over the screw axis orientation vector $\lhat$ and the direction of moment vector $\hat{\m}$ using two independent von Mises-Fisher distributions and imposes a soft orthogonality constraint over the modes of the two distributions. The distributions over the moment vector magnitude $\norm{\m}$ and configurations $q_{1:n-1}$ are considered to be normal distributions. This method suffers from the same drawback as ScrewNet, i.e., the use of a soft orthogonality constraint during training, and therefore cannot predict a valid set of screw axis parameters directly, unlike DUST-net. 
The second baseline method (Direct $F$) uses the same probability distribution as DUST-net to represent the uncertainty over the articulation model parameters, but estimates the individual elements of the \textbf{F} matrix directly. As a result, it fails to capture the uncertainty over model parameters accurately.

\begin{figure}
    \centering
    \includegraphics[width=0.85\linewidth, height=0.33\linewidth]{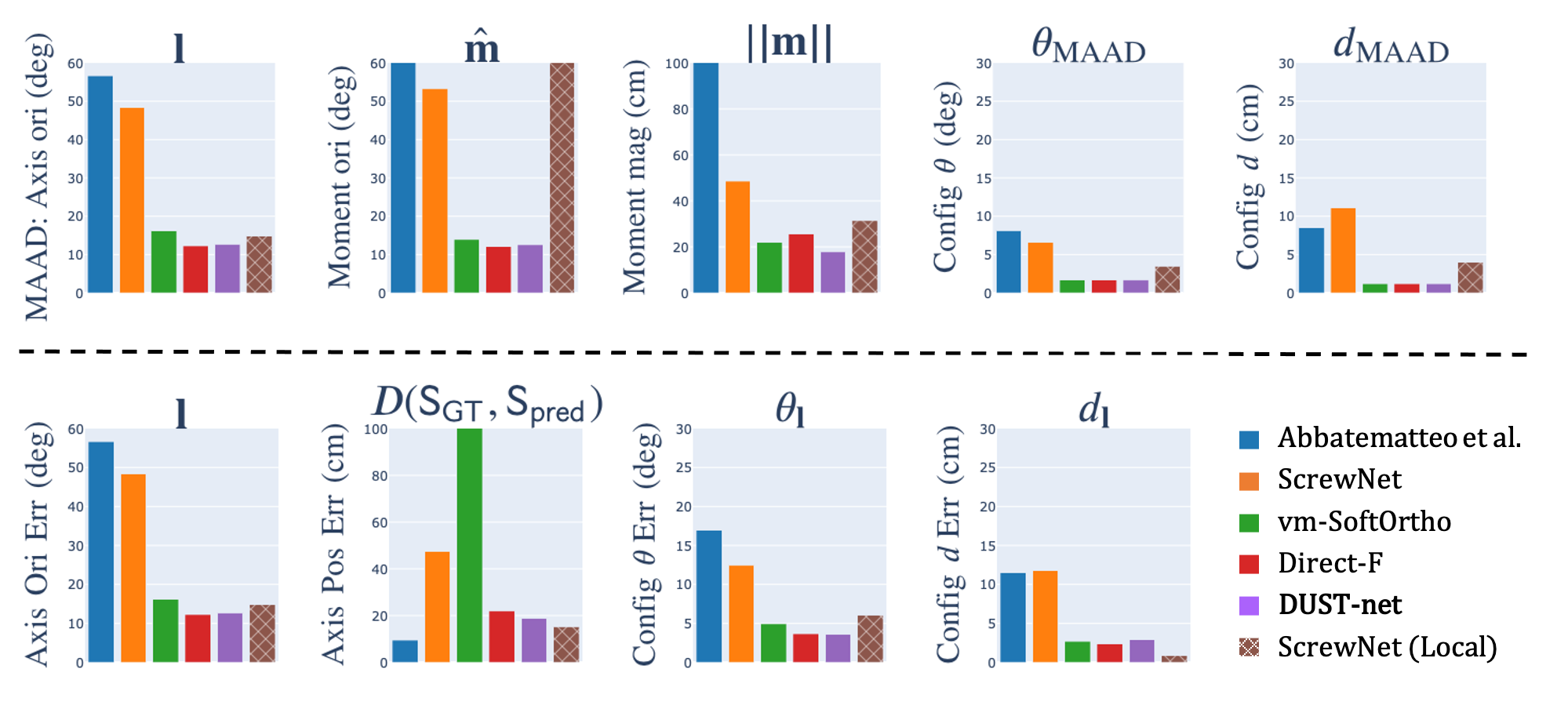}
    \caption{
    Mean error values on the MAAD (top) and Screw Loss (bottom) metrics for the PartNet-Mobility dataset \cite{Xiang_2020_SAPIEN, Mo_2019_CVPR, chang2015shapenet} (lower values are better). 
    Point estimates for DUST-net (violet) correspond to the modes of the distributions predicted by DUST-net.
    }
    \label{fig:pt_partnet}
\end{figure}

\subsection{Accuracy of Point Estimates}
The first set of experiments evaluated DUST-net’s accuracy in predicting point estimates for articulation model parameters. We use the mode of the estimated distribution as the point estimate for model parameters. We used two metrics to evaluate accuracy: Mean Absolute (Angular) Deviation (MAAD) 
and Screw Loss (Metric proposed in ScrewNet \citep{jain2020screwnet}). 
MAAD metric indicates how close the individual screw parameters are to targets, whereas the Screw Loss indicates how close the complete predicted screw transforms is to the target transforms.
The MAAD metric calculates the angular distance between the orientation of the predicted and ground-truth axis orientation vectors $\lhat$ and the orientation vectors of the screw axis moment vectors $\hat{\m}$. For the remaining parameters ($\norm{\m}, \theta_{1:n-1}, d_{1:n-1}$), it calculates the mean absolute deviation between the predicted and ground-truth values. 
The screw loss reports the angular distance between the predicted and ground-truth screw axis orientation vectors $\lhat$ as orientation error and the length of the shortest perpendicular between the predicted and ground-truth screw axes as the distance between them. Configuration errors $\theta_{1:n-1}$ are reported as the difference between the predicted rotation about the predicted screw axis and the true rotation, whereas errors over $d_{1:n-1}$ are calculated as the Euclidean distance between the points displaced by the predicted and true displacements along respective axes. 

Results for the synthetic articulated objects dataset and the PartNet-Mobility dataset are shown in Figures~\ref{fig:pt_synart} and ~\ref{fig:pt_partnet}, respectively. Results demonstrate that under both metrics, the estimates obtained from DUST-net are typically more accurate than those obtained from the state-of-the-art methods. 
The first baseline, vm-SoftOrtho, performs comparably with DUST-net on both datasets when only MAAD estimates are considered. However, Figures~\ref{fig:pt_synart} and \ref{fig:pt_partnet} show that it produces a very high distance ($\approx 1$m) between the predicted and ground-truth screw axes. This error arises due to the soft-orthogonality constraint used by vm-SoftOrtho, as DUST-net and the second baseline method, both of which handle the constraint implicitly, do not report high errors on that metric. Meanwhile, the second baseline, Direct $F$, performs comparably with DUST-net on both metrics for both datasets, but fails to capture the uncertainty over parameters with the required accuracy.

\begin{wrapfigure}{R}{0.5\textwidth}
    \centering
    \includegraphics[width=0.48\textwidth]{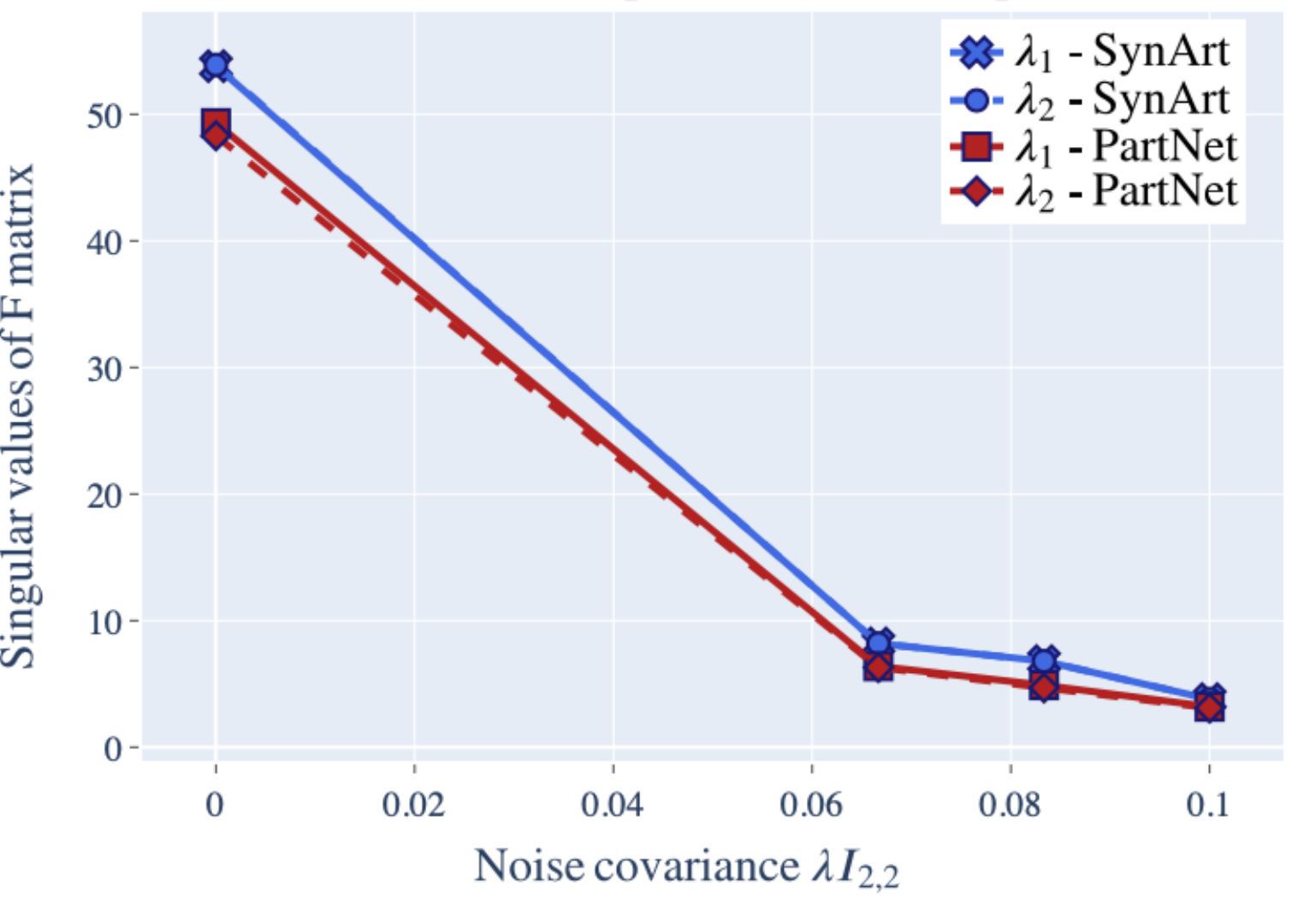}
    \caption{Variation of the mean of the singular values of predicted distribution concentration matrices over screw axes by DUST-net with artificially injected noise. Predicted singular values decrease monotonically with input noise, showing that the network’s confidence over the predicted parameters decreases with input noise.}
    \label{fig:uncertainty}
\end{wrapfigure}

\subsection{Uncertainty Estimation}
The second set of experiments evaluated how effectively DUST-net's predicted distribution captures epistemic uncertainty over the predicted articulation parameters. We evaluate this by adding artificial noise to the training labels from the two simulated datasets while training DUST-net. As more noise is added, we expect the confidence estimates produces by DUST-net to decrease as well. We add noise to the labels by sampling perturbations from a matrix von Mises-Fisher distribution with varying singular values $\lambda_1$ and $\lambda_2$ of the distribution parameter matrix $\mathbf{F}$ and the truncated normal distributions with varying precision parameters $\beta_{j}, j\in \{\norm{\m}, \theta, d\}$. Figure~\ref{fig:uncertainty} show the variation of the mean of the singular values of the predicted distribution concentration matrices over screw axes by DUST-net with injected noise. In the noiseless case, the singular values of the matrix von Mises-Fisher distribution increases until they reach their maximum allowed value at $\lambda_{max}=50$. When label noise is added, our results show that DUST-net's confidence over its predicted parameters decreases monotonically as more noise is added to the labels, supporting our hypothesis.

\begin{figure}[t]
    \centering
    \includegraphics[width=\linewidth, height=0.3\linewidth]{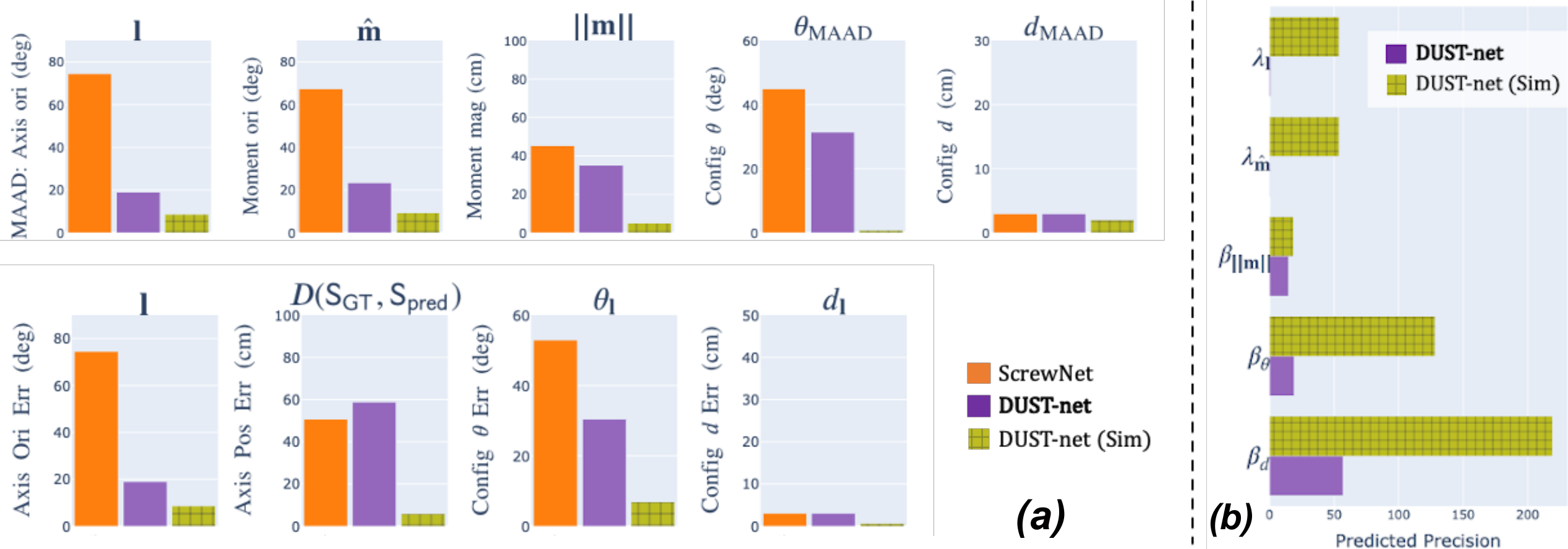}
    \caption{\textbf{(a)} Mean error values on MAAD (top) and Screw Loss (Bottom) metrics for real-world objects when the network was trained solely using simulated data~\citep{abbatematteo2019learning} (lower values are better) \textbf{(b)} Predicted concentrations over articulation model parameters. DUST-net estimation performance on simulated data~\citep{abbatematteo2019learning} (hatched green) included for comparison. DUST-net reported lower confidence in its predictions for real-world objects than simulated data (b), analogous to its degraded estimation accuracy(a).}
    \label{fig:pt_real}
\end{figure}

\subsection{Sim to Real Transfer}
\label{sec:sim2real}
Lastly, we evaluated how effectively DUST-net transfers from simulation to a real-world setting. DUST-net was trained solely on the simulated articulated object dataset~\citep{abbatematteo2019learning}. Afterward, we used it to infer the articulation model parameters for three real-world objects. 
Results (Fig.~\ref{fig:pt_real}(a)) report that DUST-net outperforms the current state-of-the-art method, ScrewNet, in estimating the model parameters for real-world objects. However, the estimated parameters using DUST-net are not yet accurate enough to be used directly for manipulating these objects. 
This sub-par performance stems from the significant differences between the training (clean and information-rich simulation data) and test datasets, which consists of noisy depth images acquired with a Kinect sensor and contain high salt-and-pepper noise, spurious features, and incomplete objects. Better performances could be achieved by either fine-tuning the network on real-world data or retraining it using a larger real-world dataset.
A noteworthy insight from the results is that DUST-net also reports low confidence over the predicted parameters for real-world objects, compared to when tested on the simulated data (Fig.~\ref{fig:pt_real}(b)). This clearly delineates why it is beneficial to estimate a distribution over the articulation model parameters instead of point estimates. Given only point estimates of articulation model parameters, a robot has no way to determine if the estimates are reliable for manipulating the object safely or not. In contrast, DUST-net's reported confidence over the predictions could allow the robot to develop safe motion policies for articulated objects~\citep{jain2018efficient, taylor20safe}
or use active learning based methods~\citep{cui2018active} to reduce uncertainty over the articulation parameters.


\section{Conclusion}
We introduced DUST-net, which utilizes a novel distribution over screw transforms on a Stiefel manifold to perform category-independent articulation model estimation with uncertainty estimates. We evaluated our approach on two benchmarking datasets and three real-world objects and compared its performance with two current state-of-the-art methods~\citep{jain2020screwnet, abbatematteo2019learning}. Results show that DUST-net can estimate articulation models, their parameters, and model uncertainty estimates for novel objects across articulation model categories successfully with better accuracy than the state-of-the-art methods. 
At present, DUST-net can only predict parameters for 1-DOF articulation models directly. For multi-DoF objects, an additional image segmentation step is required to mask out all non-relevant object parts. This procedure can be repeated iteratively for all object part pairs to estimate relative models between object parts that can be combined later to construct a complete kinematic model for the object~\citep{jain2019learning}. An interesting extension of DUST-net could estimate parameters for multi-DoF objects directly by learning a segmentation network along with it. Another exciting direction of future work is to use DUST-net in an active learning setting where, if the robot is not confident enough about the estimates of the articulation model parameters, it can actively take information-gathering actions to reduce uncertainty.

\acknowledgments{This work has taken place in the Personal Autonomous Robotics Lab (PeARL) at The University of Texas at Austin. PeARL research is supported in part by the NSF (IIS-1724157, IIS-1638107, IIS-1749204, IIS-1925082), ONR (N00014-18-2243), AFOSR (FA9550-20-1-0077), and ARO (78372-CS).  This research was also sponsored by the Army Research Office under Cooperative Agreement Number W911NF-19-2-0333. The views and conclusions contained in this document are those of the authors and should not be interpreted as representing the official policies, either expressed or implied, of the Army Research Office or the U.S. Government. The U.S. Government is authorized to reproduce and distribute reprints for Government purposes notwithstanding any copyright notation herein.}


\bibliography{references}  



\clearpage
\appendix

\section{Mathematical Background}
\label{sec:app_background}
DUST-net uses a reparameterization of the space of rigid body transformations that allows distributions over an object's articulation model parameters to be defined naturally. Here, we briefly describe the mathematical foundation leveraged in the proposed distribution over articulation parameters.

\subsection{Screw Transformations} Chasles' theorem states that \textit{``Any displacement of a body in space can be accomplished by means of a rotation of the body about a unique line in space accompanied by a translation of the body parallel to that line"}~\citep{siciliano2016springer}. Such a line is called a screw axis, $\mathsf{S}$. We represent this line using Pl\"{u}cker coordinates, given as $(\lhat, \m)$ for a $l = \mathbf{p} + x\lhat$, with moment vector $\mathbf{m} = \mathbf{p} \times \lhat$,~\citep{siciliano2016springer, jia2019}. The constraints $\norm{\lhat} = 1$ and $\langle\lhat,\m\rangle = 0$ ensure that the degrees of freedom of the line in space are restricted to four. The rigid body displacement in $SE(3)$ as a screw transform is then defined as $ \sigma = (\mathbf{l}, \mathbf{m}, \theta, d)$, where the linear displacement $d$ and the rotation $\theta$ are connected through the pitch $h$ of the screw axis, $d = h \theta$. 

\subsection{Stiefel manifold:} The \textit{Stiefel manifold} $\mathrm{V}_{k, m}$ is the space whose points are sets of $k$ orthonormal vectors in $\mathbb{R}^m$, called $k$-frames in $\mathbb{R}^m~(k \leq m)$~\citep{chikuse2003statistics}. Points on the Stiefel manifold $\mathrm{V}_{k, m}$ are represented by the set of $m \times k$ matrices $X$ such that $X^T X = I_k$, where $I_k$ is the $k \times k$ identity matrix; thus $\mathrm{V}_{k, m} = \{X_{m, k}; X^TX = I_k \}$. Some special cases of the Stiefel manifold are the unit hypersphere $\mathrm{V}_{1, m}$ in $\mathbb{R}^m$ for $k=1$, and the orthogonal group $O(m)$ 
for $m=k$. 

\subsection{von Mises-Fisher distribution} 
\begin{figure}[b]
     \centering
     \begin{subfigure}[b]{0.3\textwidth}
         \centering
         \includegraphics[width=\textwidth]{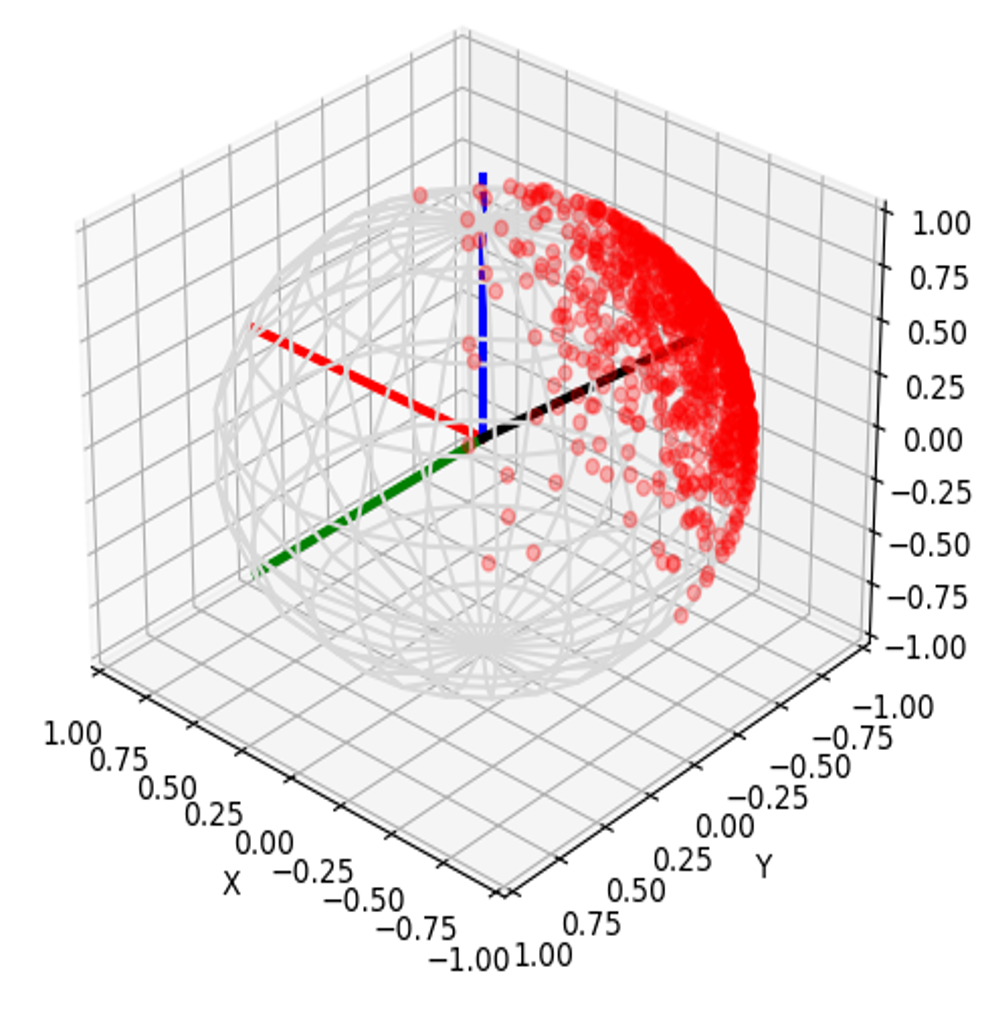}
         \caption{von Mises-Fisher distribution in $\mathbb{R}^3$. X, Y, Z axes are shown in red, blue and green colors, respectively. Black color represents the mean direction of distribution}
         \label{fig:vm-fisher}
     \end{subfigure}
     \hfill
     \begin{subfigure}[b]{0.6\textwidth}
         \centering
         \includegraphics[width=\textwidth]{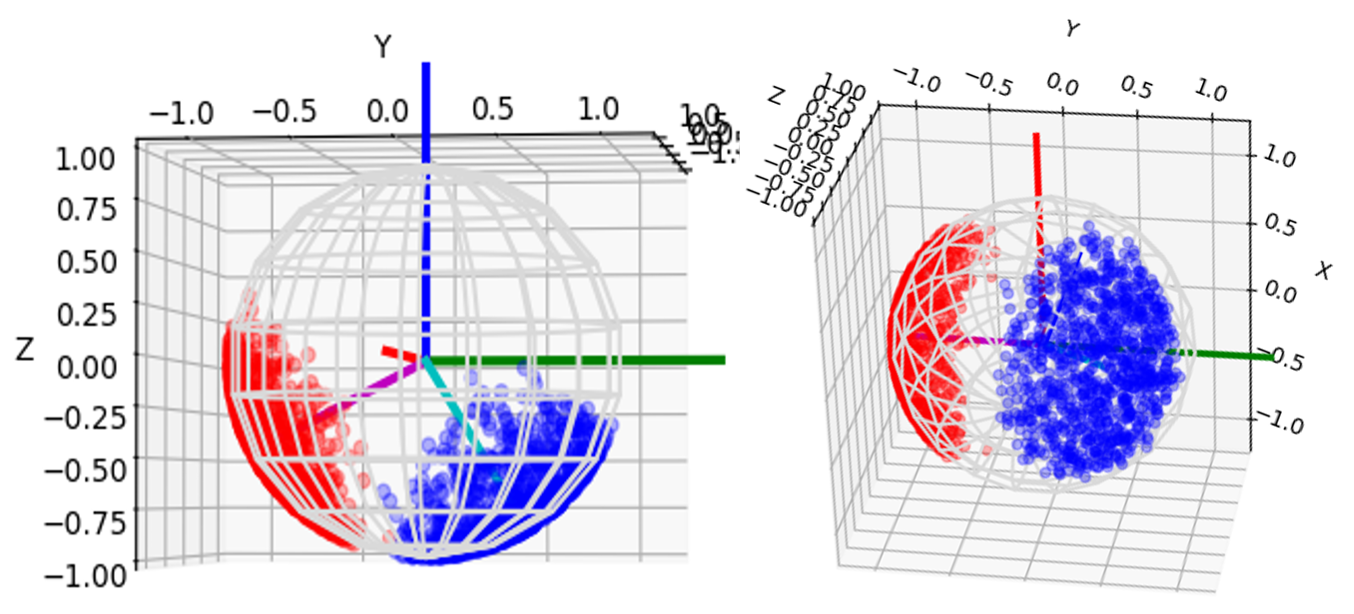}
         \caption{Matrix von Mises-Fisher distribution over $\mathrm{V}_{3,2}$,  X, Y, Z axes are shown in red, blue and green colors, respectively. Magenta and cyan colors denote vectors corresponding to the first and second column of the matrix $M \in \mathrm{V}_{3,2}$ representing the mode of the distribution}
         \label{fig:vmSt}
     \end{subfigure}
\end{figure}

The von Mises-Fisher distribution (or Langevin distribution) is a unimodal probability distribution on the $(m-1)$ sphere in $\mathbb{R}^m$ (see Figure~\ref{fig:vm-fisher}). A random $m$-dimensional unit vector $\bm{x}$ is said to have the von Mises–Fisher distribution, if its probability distribution function is given by: $f_m (\bm{x} | \boldsymbol{\mu}, \kappa) = C_m(\kappa) \exp(\kappa \boldsymbol{\mu}^T\bm{x})$, where the concentration parameter $\kappa \geq 0 $, the mean direction $\norm{\boldsymbol{\mu}} = 1$ and the normalization constant $C_m(\kappa) = \dfrac{\kappa^{\frac{m}{2} - 1}}{(2 \pi)^{\frac{m}{2}} I_{\frac{m}{2} - 1} (\kappa)}$ where $I_{\nu}$ denotes the modified Bessel function of the first kind at order $\nu$~\citep{mardia1999directional}. For $m=3$, the normalization constant reduces to $C_3(\kappa) = \dfrac{\kappa}{4 \pi \sinh \kappa} = \dfrac{\kappa~e^{-\kappa}}{2 \pi (1 - e^{-2\kappa})}$.

\subsection{Matrix von Mises-Fisher distribution} A random matrix $X$ on $\mathrm{V}_{k, m}$ is said to have the matrix von Mises-Fisher distribution (or matrix Langevin distribution), if its density function is given by $\mathcal{F}(\mathbf{X}| m , \mathbf{F}) = \dfrac{1}{_0F_1 (\frac{m}{2}, \frac{1}{4} \mathbf{F}^T \mathbf{F})}~\exp(\mathrm{Tr}(\mathbf{F}^T \mathbf{X}))$, where \textbf{F} is any $m\times k$ matrix and $_0F_1$ is a hypergeometric function with matrix argument~\citep{chikuse2003statistics} (see Figure~\ref{fig:vmSt} for an illustration). We can write the general (unique) singular value decomposition (SVD) of \textbf{F} as $\bm{F} = \Gamma \Lambda \Omega^T$, where $\Gamma \in \tilde{\mathrm{V}}_{k, m}$, $\Omega \in O(k)$, $\Lambda = diag(\lambda_1,...,\lambda_k),~~\lambda_1 \geq ... \geq \lambda_k \geq 0$, $\tilde{\mathrm{V}}_{k, m}$ denotes the set of matrices $\Gamma \in \mathrm{V}_{k, m}$ with the property that all the elements of the first row of the matrix $\Gamma$ are positive, and $O(k)$ denoting the orthogonal group in $k$ dimensions. It can be shown that $~_0F_1 (\frac{m}{2}, \frac{1}{4} \mathbf{F}^T \mathbf{F}) =~_0F_1 (\frac{m}{2}, \frac{1}{4} \Lambda^2)$. For more details, we refer to~\citep{chikuse2003statistics}.

\section{Joint distribution over model parameters}
\label{sec:app_model}
A screw transform, represented as a tuple $\langle \mathsf{S}, \theta, d \rangle$, corresponds to a point on the manifold $\mathbb{S} \times SO(2) \times \mathbb{R}^+$, where $\mathbb{S} := \mathrm{V}_{2, 3} \times \mathbb{R}^+ $, $\mathrm{V}_{2, 3}$ is the Stiefel manifold of $2-$frames in $\mathbb{R}^3$, $SO(2)$ denotes the circle group or the special orthogonal in two dimensions, and $\mathbb{R}^+$ denotes the set of positive real numbers. The unified representation proposed by \citet{jain2020screwnet} considers the motion of an articulated object as a sequence of screw transforms that share a common screw axis $\mathsf{S}$. Hence, the extended tuple $\langle \mathsf{S}, \theta_{1:n-1}, d_{1:n-1} \rangle$, representing the articulation model for an object, corresponds to a point on the manifold $\mathbb{S} \times [SO(2)]^{n-1} \times [\mathbb{R}^+]^{n-1}$.  
We can define a joint distribution over the articulation model parameters by defining the probability density function for the distribution as the exponentiated distance of a point from the modal point of the distribution, and subsequently restricting the density function to the manifold~\citep{chikuse2003statistics}. However, calculating the normalization factor for this distribution is challenging.
For example, a direct extension of the von Mises-Fisher distribution to define a distribution on $\mathrm{V}_{2, 3} \times \mathbb{R}$ yields a density function with a normalizing factor that requires integrating a generalized hypergeometric function, which, to the best of our knowledge, is not computationally tractable to compute \cite{khatri1977mises,james1964distributions}.
Therefore, to define a distribution over the articulation model parameters that is tractable to learn, we make certain assumptions and propose an approximate joint distribution over the model parameters in this work.

Given a sequence of $n$ depth images $\mathcal{I}_{1:n}$ of object part motion, the joint probability distribution over the articulation model parameters $p(\mathsf{S}, \theta_{1:n-1}, d_{1:n-1}~|~\mathcal{I}_{1:n})$ can be written as a product of a distribution over the screw axis parameters and a conditional distribution over the joint configuration parameters:
\begin{equation}
    p(\mathsf{S}, \theta_{1:n-1}, d_{1:n-1}~|~\mathcal{I}_{1:n}) = p(\mathsf{S}~|~\mathcal{I}_{1:n})~p(\theta_{1:n-1}, d_{1:n-1} ~|~\mathsf{S},\mathcal{I}_{1:n})
    \label{eq:app_jnt}
\end{equation}
We first approximate the distribution over the screw axis parameters $\mathsf{S}$ as a product of two marginal distributions: one over the orientation vector tuple $\langle \lhat, \hat{\m} \rangle \in \mathrm{V}_{2,3}$ and another over the moment vector magnitude $\norm{\m} \in \mathbb{R}^+$, 
\begin{equation}
    p(\mathsf{S}~|~\mathcal{I}_{1:n}) \approx p(\langle \lhat, \hat{\m} \rangle ~|~\mathcal{I}_{1:n}) ~ p(\norm{\m} ~|~\mathcal{I}_{1:n})
    \label{eq:app_screw}
\end{equation}
This approximation is motivated by the fact that calculating statistics over manifolds can be computationally intractable in a general setting~\citep{chikuse2003statistics, mardia1999directional, jiu2020calculation}. This approximation enables us to define the probability density function over the screw axis parameters using standard distributions over manifolds whose properties are well studied in the literature, such as the matrix von Mises-Fisher distributions over Stiefel manifolds~\citep{chikuse2003statistics, mardia1999directional}.

Calculating the conditional distribution over joint configurations, $p(\theta_{1:n-1}, d_{1:n-1} ~|~\mathsf{S},\mathcal{I}_{1:n})$, exactly would require us to evaluate hypergeometric functions over the complete manifold in which the screw transforms lie. Hypergeometric functions in the matrix argument result in an infinite series in terms of zonal polynomials, which becomes combinatorially expensive to calculate with the increasing number of terms~\citep{jiu2020calculation}. To maintain the numerical tractability of the solution, we approximate the probability density function of the conditional distribution as a Dirac delta function centered at the expected value of the distribution over the screw axis parameters $\bar{\mathsf{S}}$:
\begin{align}
    \begin{split}
        p(\theta_{1:n-1}, d_{1:n-1} ~|~\mathsf{S},\mathcal{I}_{1:n}) \approx ~& \delta_{\bar{\mathsf{S}}}[p(\theta_{1:n-1}, d_{1:n-1} ~|~\mathsf{S},\mathcal{I}_{1:n})] \\
        = ~& p(\theta_{1:n-1}, d_{1:n-1} ~|~\bar{\mathsf{S}},\mathcal{I}_{1:n})
    \end{split}
    \label{eq:app_conditional}
\end{align}
where $\bar{\mathsf{S}} = \int_{\mathbb{S}} \mathsf{S}~p(\mathsf{S}~|~\mathcal{I}_{1:n})$. 


As we noted earlier, the unified parameterization of the articulation model parameters corresponds to a sequence of rigid body transforms (or screw transforms). Each of these rigid body transforms can be treated as an independent frame transformation between the object parts. Leveraging this fact, we approximate the conditional distribution over the joint configurations as a product of marginals over screw transforms at each time step:
\begin{align}
    \begin{split}
        p(\theta_{1:n-1}, d_{1:n-1} ~|~\bar{\mathsf{S}},\mathcal{I}_{1:n}) &= \prod^{n-1}_{i=1} p(\theta_{i}, d_{i} ~|~\bar{\mathsf{S}},\mathcal{I}_{1:n})
    \end{split}
    \label{eq:app_seq}
\end{align}

In this work, we approximate the conditional distribution over the joint configurations, $p(\theta_{i}, d_{i} ~|~\bar{\mathsf{S}},\mathcal{I}_{1:n})$, as a product of marginals over the rotation and displacement parameters to further simplify the parameterization of the joint distribution over articulation model parameters:
\begin{equation}
    p(\theta_{i}, d_{i} ~|~\bar{\mathsf{S}},\mathcal{I}_{1:n}) \approx p(\theta_{i} ~|~\bar{\mathsf{S}},\mathcal{I}_{1:n}) ~ p(d_{i} ~|~\bar{\mathsf{S}},\mathcal{I}_{1:n})
    \label{eq:app_conf}
\end{equation}
While this approximate distribution cannot capture the correlations between joint configurations, it was found to be sufficiently expressive to enable DUST-Net to outperform the state-of-the-methods for articulation model estimation with a significant margin (see Section~\ref{sec:results}). In the future, DUST-Net may be extended to use multivariate distributions instead, which can capture the correlations between joint configurations as well. 

Combining these together, in this work, we propose to approximate the joint distribution over articulation model parameters as:
\begin{align}
    \begin{split}
        p(\mathsf{S}, \theta_{1:n-1}, d_{1:n-1}~|~\mathcal{I}_{1:n}) &\approx p(\mathsf{S}~|~\mathcal{I}_{1:n})~\prod^{n-1}_{i=1} p(\theta_{i}~|~\bar{\mathsf{S}},\mathcal{I}_{1:n})~ \prod^{n-1}_{i=1} p(d_{i} ~|~\bar{\mathsf{S}},\mathcal{I}_{1:n}) \\ 
        &\approx p(\langle \lhat, \hat{\m} \rangle | \mathcal{I}_{1:n}) ~ p(\norm{\m} | \mathcal{I}_{1:n})
        ~\prod^{n-1}_{i=1} p(\theta_{i}~|~\bar{\mathsf{S}},\mathcal{I}_{1:n})~ \prod^{n-1}_{i=1} p(d_{i} ~|~\bar{\mathsf{S}},\mathcal{I}_{1:n})
    \end{split}
\end{align}
where the exact parameterization of each of these probability distribution functions is discussed in section~\ref{sec:approach} of the main text.



\section{Hypergeometric function \texorpdfstring{${}_pF_{q}$}{qFp} }
\label{app:zonal}
A general hypergeometric function ${}_pF_{q}$ in the matrix argument can be written as an infinite series in terms of zonal polynomials, which are multivariate symmetric homogeneous polynomials and form a basis of the space of symmetric polynomials~\citep{chikuse2003statistics}. Given an $m\times m$ symmetric, positive-definite matrix Y, the hypergeometric function ${}_pF_{q}$ of matrix argument Y is defined as
\begin{equation}\label{eq:MatrixpFq}
  {}_{p}F_{q}\left(\genfrac{}{}{0pt}{}{a_{1},\ldots,a_{p}}{b_{1},\ldots,b_{q}}\,\bigg|\,Y\right) :=
  \sum_{n=0}^{\infty}\sum_{\nu\in\mathcal{P}_{n}}\frac{(a_{1})_{\nu}\cdots(a_{p})_{\nu}}
      {(b_{1})_{\nu}\cdots(b_{q})_{\nu}}\cdot\frac{\mathcal{C}_{\nu}(Y)}{n!},
\end{equation}
where 
\begin{itemize}
\item $\mathcal{P}_{n}$ is the set of all ordered integer partitions of $n$
\item $(a)_{\nu}$ is the generalized Pochhammer symbol, defined as
\[
(a)_{\nu}=(a)_{(\nu_{1},\dots,\nu_{k})}:=\prod_{i=1}^{k}\left(a-\frac{i-1}{2}\right)_{\!\!\nu_{i}}\!\!;
\],
where, $(a)_{\nu_i} = a(a+1)...(a + \nu_i - 1), (a)_0 = 1$,
\item and $\mathcal{C}_{\nu}(Y)$ denotes the zonal polynomial
of~$Y$, indexed by a partition~$\nu$, which is a symmetric homogeneous polynomial of degree~$n$ in the eigenvalues $y_{1},\ldots,y_{m}$ of~$Y$, satisfying 
\begin{equation}\label{eq:TrZonal}
  \sum_{\nu\in\mathcal{P}_{n}}\mathcal{C}_{\nu}(Y)=(\Tr Y)^{n}=(y_{1}+\cdots+y_{m})^{n}.
\end{equation}
\end{itemize}

Using zonal polynomials, we can define the hypergeometric function ${}_0F_{1} (\frac{3}{2}, \frac{1}{4} \Lambda^2)$ defining the normalization factor of the matrix von Mises-Fisher distribution over Stiefel manifold $\mathcal{V}_{3,2}$ as
\begin{equation}
    {}_0F_{1} (\frac{3}{2}, \frac{1}{4} \Lambda^2) := \sum^\infty_{n=0} \sum_{\nu \in \mathcal{P}_n} \dfrac{1}{(\frac{3}{2})_{\nu}} \dfrac{C_{\nu}(\Lambda)}{n!},  
\end{equation}
where $\Lambda = \diag(\lambda_1, \lambda_2)$, $\mathcal{P}_n$ is the set of all ordered integer partitions of $n$, $(a)_\nu$ is the generalized Pochhammer symbol, and $C_\nu (\Lambda)$ denotes the zonal polynomial of $\Lambda$ indexed by a partition $\nu$. This series converges for all input matrices for a general hypergeometric function ${}_pF_{q}$ if $p \leq q$, which holds in our case \citep{chikuse2003statistics}. Recently, \citet{jiu2020calculation} investigated the zonal polynomials in detail and developed a computer algebra package to calculate these polynomials in SageMath. We use this package to calculate the the hypergeometric function ${}_0F_{1} (\frac{3}{2}, \frac{1}{4} \Lambda^2)$. However, as the number of terms in the series grows combinatorially with $n$, we truncate the series at $n=25$ for computational reasons. Through our experimental analysis, we found that this truncated series is a good approximation of ${}_0F_{1}$ as the series converges to a finite value, if the singular values of the $F$, i.e. $\lambda_1$ and $\lambda_2$ remain below a maximum value $\lambda_{max}=50$.

\begin{figure}
    \centering
    \includegraphics[width=0.8\linewidth]{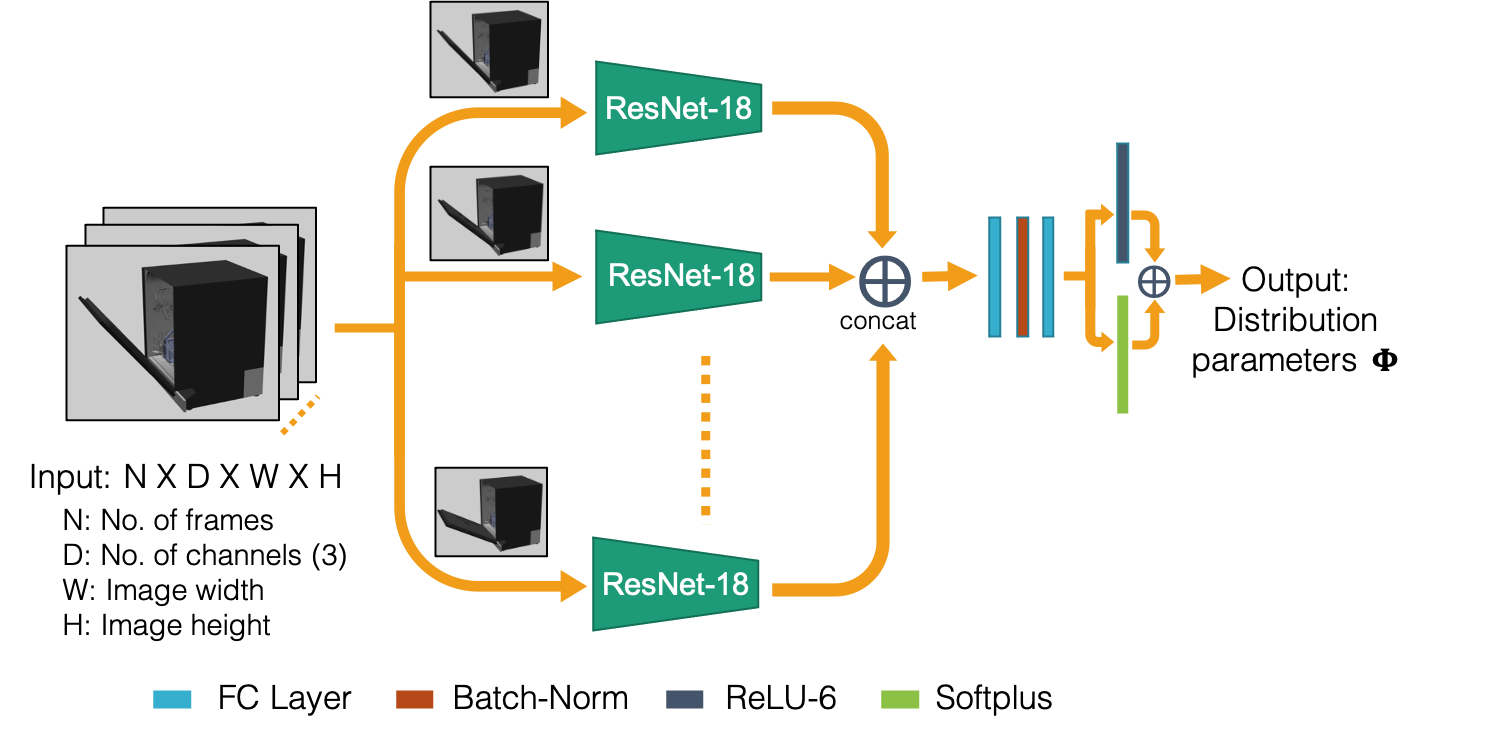}
    \caption{DUST-net architecture}
    \label{fig:arch}
\end{figure}

\section{Network Architecture}
Figure~\ref{fig:arch} shows the detailed network architecture for DUST-net. DUST-net uses an off-the-shelf convolutional network, ResNet-18, to extract task-relevant visual features from the input images, which are later passed through a two-layer MLP to predict a set of parameters $\Phi$ for the distribution $p(\mathsf{S}, \theta_{1:n-1}, d_{1:n-1} ~|~ \mathcal{I}_{1:n}, \Phi)$. We use ReLU activations for the hidden fully-connected layers. The first four output parameters (out of 40) of the last linear layer of MLP correspond to the parameters $(\alpha, \beta, \gamma)$ and $\omega$, representing the matrices $\Gamma$ and $\Omega$ respectively, which lie in ranges $[0, 2\pi), [0, \pi), [0, 2\pi)$, and $[0, 2\pi)$ respectively. We pass the first four values of the output of the last linear layer through a ReLU-6 layer~\citep{howard2017mobilenets} to correctly map the predicted values with their respective ranges. The rest of the parameters are required to be non-negative. We pass the remaining output values of the last linear layer through a Softplus layer for non-negative output. 


 
\begin{figure}[b]
    \centering
    \includegraphics[width=0.8\linewidth]{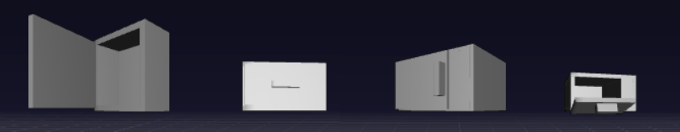}
    \caption{Object classes used from the simulated articulated object dataset \cite{abbatematteo2019learning}. Object classes: cabinet, drawer, microwave, and toaster (left to right)}
    \label{fig:simArt-dataset}
\end{figure}

\section{Experimental details}
\subsection{Datasets}
\begin{figure}[t]
    \centering
    \includegraphics[width=0.8\linewidth]{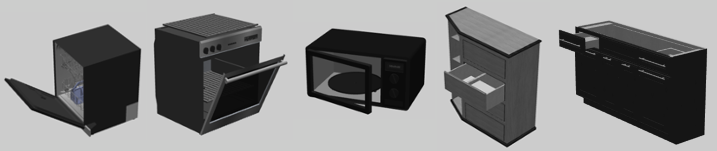}
    \caption{Object classes used from the PartNet-Mobility dataset \cite{Xiang_2020_SAPIEN, Mo_2019_CVPR, chang2015shapenet}. Object classes: dishwasher, oven, microwave, drawer- 1 column, and drawer- multiple columns (left to right)}
    \label{fig:partNet-dataset}
\end{figure}

\begin{figure}[t]
    \centering
    \includegraphics[width=0.6\linewidth]{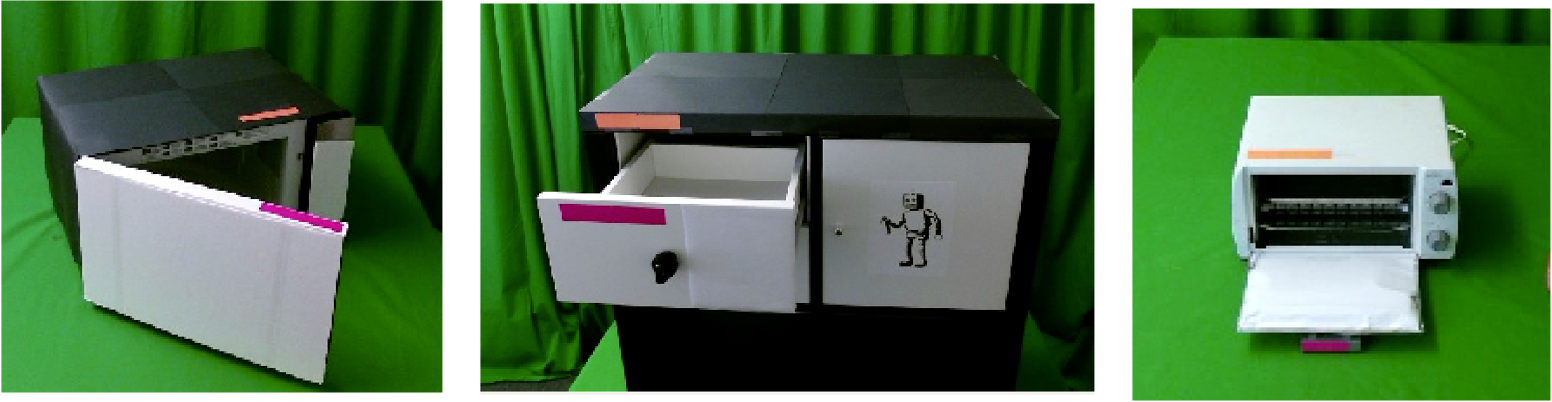}
    \caption{Real world objects used to evaluate DUST-net's performance. Object classes: microwave, drawer, and toaster (left to right)}
    \label{fig:real-world-dataset}
\end{figure}

Objects used in the experiments from each of the dataset are shown in the Figures~\ref{fig:simArt-dataset} and \ref{fig:partNet-dataset}. We sampled a new object geometry and a joint location for each training example in the simulated articulated object dataset, as proposed by \cite{abbatematteo2019learning}. For the PartNet-Mobility dataset, we considered $11$ microwave ($8$ train, $3$ test), $36$ dishwasher ($27$ train, $9$ test), $9$ oven ($6$ train, $3$ test), $26$ single column drawer ($20$ train, $6$ test), and $14$ multi-column drawer ($10$ train, $4$ test) object models. For both datasets, we sampled object positions and orientations uniformly in the view frustum of the camera up to a maximum depth dependent upon the object size. The objects and depth images are rendered in Mujoco \cite{todorov2012mujoco}. We apply random frame skipping and pixel dropping to simulate noise encountered in real world sensor data. We consider three household objects --- a microwave, a drawer, and a toaster oven, in the real world objects dataset for evaluating DUST-net's performance. The objects are shown in Figure~\ref{fig:real-world-dataset}. 

To generate the labels for screw displacements, we follow the same procedure as used by~\citet{jain2020screwnet}. Considering one of the objects, $o_i$, as the base object, we calculate the screw displacements between temporally displaced poses of the second object $o_j$ with respect to it. Given a sequence of $n$ images $\mathcal{I}_{1:n}$, we calculate a sequence of $n-1$ screw displacements $^1\boldsymbol{\sigma}_{o_j} =\{^1\sigma_2,...^1\sigma_{n} \}$, where each $^1\sigma_k$ corresponds to the relative spatial displacement between the pose of the object $o_j$ in the first image $\mathcal{I}_1$ and the images $\mathcal{I}_{k,~k\in \{2...n\}}$. Note $^1\boldsymbol{\sigma}_{o_j}$ is defined in the frame $\mathcal{F}_{o_j^1}$ attached to the pose of the object $o_j$ in the first image $\mathcal{I}_1$. We then transform $^1\boldsymbol{\sigma}_{o_j}$ to the camera frame by defining the 3D line motion matrix $\tilde{D}$ between the frames $\mathcal{F}_{o^1_j}$ and $\mathcal{F}_{o_i}$ \cite{bartoli20013d}, and transforming the common screw axis $\mathsf{^1S}$ to the target frame $\mathcal{F}_{o_i}$. The configurations $^1q_k$ remain the same during frame transformations. The 3D line motion matrix $\tilde{D}$ between two frames can be constructed using the rotation matrix $R$ and a translation vector $\mathbf{t}$ between two frames $\mathcal{F}_A$ and $\mathcal{F}_B$, as: 
\begin{equation}
    \begin{gathered}
        \begin{bmatrix}
        ^B\lhat \\ ^B\m
        \end{bmatrix} = ~^B\tilde{D}_A ~ \begin{bmatrix} ^A\lhat \\ ^A\m \end{bmatrix}, ~~~~~~ \text{where,} ^B\tilde{D}_A = 
        \begin{bmatrix}
            R & \mathbf{0} \\
            [\mathbf{t}]_{\times}R & R
        \end{bmatrix},
        [\mathbf{t}]_{\times} = 
        \begin{bmatrix}
            0 & -t_3 & t_2 \\
            t_3 & 0 & -t_1 \\
            -t_2 & t_1 & 0
        \end{bmatrix}
        \label{eq:2}
    \end{gathered}
\end{equation}
where $[\mathbf{t}]_{\times}$ denotes the skew-symmetric matrix corresponding to the translation vector $\mathbf{t}$, and $(^A\lhat, ^A\m)$ and $(^B\lhat, ^B\m)$ represents the line $l$ in frames $\mathcal{F}_A$ and $\mathcal{F}_B$, respectively \cite{bartoli20013d}.

\section{Further Results}

\subsection{Accuracy of Point Estimates}
Detailed numerical results for the synthetic articulated objects dataset and the PartNet-Mobility dataset are shown in Tables~\ref{tab:synArt} and ~\ref{tab:partnet}, respectively. Results demonstrate that under both metrics, the estimates obtained from DUST-net are considerably more accurate than those obtained from the state-of-the-art methods. DUST-net also correctly estimates very high distribution concentration parameters for the true, noise-free labels. The first baseline, vm-SoftOrtho, performs comparably with DUST-net on both datasets when only MAAD estimates are considered. However, Tables~\ref{tab:synArt} and \ref{tab:partnet} show that it produces a very high distance ($\approx 1$m) between the predicted and ground-truth screw axes. This error arises due to the soft-orthogonality constraint used by vm-SoftOrtho, as DUST-net and the second baseline method, both of which handle the constraint implicitly, do not report high errors on that metric. Meanwhile, the second baseline, Direct $F$, performs comparably with DUST-net on both metrics for both datasets, but fails to capture the uncertainty over parameters with the required accuracy.

\begin{table}
\centering
\resizebox{\textwidth}{!}{%
\begin{tabular}{c|c|cc:c|c:c|c:c|ccccc}
 & MAAD / SL & \multicolumn{2}{c:}{MAAD} & Screw Loss & \multicolumn{1}{c:}{MAAD} & SL & \multicolumn{1}{c:}{MAAD} & SL & \multicolumn{4}{c}{Precision} \\
 & $\lhat$ & $\hat{\m}$ & \multicolumn{1}{c:}{$\norm{\m}$} & $D(\mathsf{S}_{GT},  \mathsf{S}_{pred})$ & \multicolumn{1}{c:}{$\theta_i$} & $\theta_{\lhat}$ & \multicolumn{1}{c:}{$d_i$} & $d_\lhat$ & $\lambda_\lhat$ & $\lambda_{\hat{\m}}$ & $\beta_{\norm{\m}}$ & $\beta_{\theta}$ & $\beta_{d}$ \\ \hline
vm-SoftOrtho & \textbf{0.139} & \textbf{0.154} & 0.068 & 0.956 & \textbf{0.012} & \textbf{0.117} & 0.003 & 0.006  & \textbf{56.2} & \textbf{55.8} & 9.8 & 47.9 & 89.5\\
Direct \textbf{F} & 0.240 & 0.261 & 0.062 & 0.104 & 0.010 & 0.208 & 0.002 & 0.006 & 8.4 & 7.9 & 9.8 & 48.5 & 75.3\\
ScrewNet & 0.846 & 0.929 & 0.486 & 0.475 & 0.115 & 0.217 & 0.111 & 0.118 & - & - & - & - & - \\
\citet{abbatematteo2019learning} & 0.194 & - & - & 0.111 & 0.223 & - & 0.045 & - & - & - & - & - & -\\
\textbf{DUST-net} & 0.151 & 0.163 & \textbf{0.052} & \textbf{0.059} & \textbf{0.012} & 0.122 & \textbf{0.002} & \textbf{0.006} & 53.8 & 54.0 & \textbf{18.3} & \textbf{128.1} & \textbf{219.1}\\
\hdashline
\rule{0pt}{3ex} ScrewNet (Local) & 0.178 & 0.443 & 0.068 & 0.033 & 0.057 & 0.118 & 0.015 & 0.015 & - & - & - & - & - \\
\end{tabular}%
}
\vspace{2pt}
\caption{Mean error values on the MAAD and Screw Loss(SL) metrics for the simulated articulated objects dataset~\citep{abbatematteo2019learning}. Point estimates for DUST-net correspond to the modes of the distributions predicted by DUST-net. Angular values \{$\lhat, \hat{\m}, \theta_i, \theta_{\lhat}$\} and distances \{$\norm{\m}, D, d_i, d_\lhat$\} are reported in radian and meter, respectively. Numerical values are reported for the uncertainty parameters \{$\lambda_i, \beta_j$\}. Symbol $-$ represents value not reported.
}
\label{tab:synArt}
\end{table}

\begin{table}
\centering
\resizebox{\textwidth}{!}{%
\begin{tabular}{c|c|cc:c|c:c|c:c|ccccc}
 & MAAD / SL & \multicolumn{2}{c:}{MAAD} & Screw Loss & \multicolumn{1}{c:}{MAAD} & SL & \multicolumn{1}{c:}{MAAD} & SL & \multicolumn{4}{c}{Precision} \\
 & $\lhat$ & $\hat{\m}$ & \multicolumn{1}{c:}{$\norm{\m}$} & $D(\mathsf{S}_{GT},  \mathsf{S}_{pred})$ & \multicolumn{1}{c:}{$\theta_i$} & $\theta_{\lhat}$ & \multicolumn{1}{c:}{$d_i$} & $d_\lhat$ & $\lambda_\lhat$ & $\lambda_{\hat{\m}}$ & $\beta_{\norm{\m}}$ & $\beta_{\theta}$ & $\beta_{d}$ \\ \hline
vm-SoftOrtho & 0.284 & 0.243 & 0.221 & 1.137 & 0.030 & 0.086 & 0.012 & 0.027  & 26.9 & 31.1 & 5.7 & 54.5 & 60.9 \\
Direct \textbf{F} & \textbf{0.214} & \textbf{0.212} & 0.257 & 0.219 & 0.030 & 0.064 & 0.012  & \textbf{0.024} & 8.1 & 7.3 & 4.9 & 59.5 & 70.9 \\
ScrewNet & 0.846 & 0.929 & 0.486 & 0.475 & 0.115 & 0.217 & 0.111 & 0.118 & - & - & - & - & - \\
\citet{abbatematteo2019learning} & 0.989 & - & - & \textbf{0.095 }& 0.141  & - & 0.085 & - & - & - & - & - & -\\
\textbf{DUST-net} & 0.220& 0.219 & \textbf{0.178} & 0.189 & \textbf{0.029} & \textbf{0.063} & \textbf{0.012} & 0.029 & \textbf{49.3} & \textbf{48.3} & \textbf{7.7} & \textbf{72.0} & \textbf{131.9} \\
\hdashline
\rule{0pt}{3ex}  ScrewNet (Local) & 0.260 & 1.23 & 0.314 & 0.151 & 0.060 & 0.106 & 0.040 & 0.009 & - & - & - & - & - \\
\end{tabular}%
}
\vspace{2pt}
\caption{Mean error values on the MAAD and Screw Loss(SL) metrics for the PartNet-Mobility dataset \cite{Xiang_2020_SAPIEN, Mo_2019_CVPR, chang2015shapenet}. Point estimates for DUST-net correspond to the modes of the distributions predicted by DUST-net. Angular values \{$\lhat, \hat{\m}, \theta_i, \theta_{\lhat}$\} and distances \{$\norm{\m}, D, d_i, d_\lhat$\} are reported in radian and meter, respectively. Numerical values are reported for the uncertainty parameters \{$\lambda_i, \beta_j$\}. Symbol $-$ represents value not reported.
}
\label{tab:partnet}
\vspace{-10pt}
\end{table}

\subsection{Uncertainty Estimation}
The detailed numerical results from the second set of experiments are shown in Table~\ref{tab:calib}. In the noiseless case, the singular values of the matrix von Mises-Fisher distribution increases until they reach their maximum allowed value at $\lambda_{max}=50$, while the precision parameters $\beta_{j}, j\in \{\norm{\m}, \theta, d\}$ for truncated normal distributions over remaining parameters become arbitrarily large. 

\begin{table}[]
\centering
\resizebox{\textwidth}{!}{%
\begin{tabular}{c|cccll|ccllc|ccllc|ccllc}
 & $\lambda_1$ & $\lambda_2$ & {$\beta_{\norm{\m}}$} & $\beta_{\theta}$ & $\beta_{d}$ & $\lambda_1$ & $\lambda_2$ & \multicolumn{1}{c}{$\beta_{\norm{\m}}$} & $\beta_{\theta}$ & $\beta_{d}$ & $\lambda_1$ & $\lambda_2$ & \multicolumn{1}{c}{$\beta_{\norm{\m}}$} & $\beta_{\theta}$ & $\beta_{d}$ & $\lambda_1$ & $\lambda_2$ & \multicolumn{1}{c}{$\beta_{\norm{\m}}$} & $\beta_{\theta}$ & $\beta_{d}$\\ \hline
Label Noise & \multicolumn{4}{c}{No noise} &  & 15 & 15 & 50 & 50 & 50 & 12 & 12 & 50 & 50 & 50 & 10 & 10 & 50 & 50 & 50 \\ \hline
SynArt & 53.8 & 53.9 & 18.3 & 128.0 & 219.0 & 8.2 & 8.2 & 14.6 & 53.7 & 51.9 & 6.8 & 6.8 & 10.5 & 41.6 & 49.6 & 3.8 & 3.8 & 10.3 & 41.9 & 47.4 \\ \hdashline
PartNet & 49.3 & 48.3 & 7.7 & 72.0 & 132.0 & 6.4 & 6.3 & 9.4 & 29.5 & 29.2 & 4.9 & 4.7 & 8.9 & 34.0 & 37.9 & 3.2 & 3.1 & 9.4 & 31.2 & 32.1
\end{tabular}%
}
\vspace{2pt}
\caption{Testing variation of DUST-net's confidence over predicted articulation model parameters with input noise. DUST-net’s confidence over its predicted parameters decreases monotonically as input noise is increased showing that DUST-net’s predicted distribution captures the network’s confidence over the predicted articulation parameters effectively.}
\label{tab:calib}
\end{table}

\subsection{Real objects}
The numerical results from the sim-to-real transfer experiments are shown in Table~\ref{tab:real-world}. Results report that while DUST-net outperforms ScrewNet in estimating the model parameters for real-world objects, the estimated parameters are not yet accurate enough to be used directly for manipulating these objects. However, a noteworthy insight from the results is that DUST-net also reported very low confidence over the predicted parameters. This clearly delineates why it is beneficial to estimate a distribution over the articulation model parameters instead of only point estimates, as discussed earlier in the section~\ref{sec:sim2real}.

\begin{table}[]
\centering
\resizebox{\textwidth}{!}{%
\begin{tabular}{l|c|c|cc:c|c:c|c:c|ccccc}
 & & MAAD / SL & \multicolumn{2}{c:}{MAAD} & Screw Loss & \multicolumn{1}{c:}{MAAD} & SL & \multicolumn{1}{c:}{MAAD} & SL & \multicolumn{4}{c}{Precision} \\
 & & $\lhat$ & $\hat{\m}$ & \multicolumn{1}{c:}{$\norm{\m}$} & $D(\mathsf{S}_{GT},  \mathsf{S}_{pred})$ & \multicolumn{1}{c:}{$\theta_i$} & $\theta_{\lhat}$ & \multicolumn{1}{c:}{$d_i$} & $d_\lhat$ & $\lambda_\lhat$ & $\lambda_{\hat{\m}}$ & $\beta_{\norm{\m}}$ & $\beta_{\theta}$ & $\beta_{d}$ \\ \hline
Toaster & ScrewNet & 2.42 & 2.48 & 0.74 & 0.76 & 0.45 & 1.26 & 0.01 & 0.00 & - & - & - & - & - \\
Oven & DUST-net & \textbf{0.17} & \textbf{0.31} & 0.\textbf{52} & \textbf{0.59} & \textbf{0.44} & \textbf{0.64} & \textbf{0.01} & \textbf{0.01} & 2.5 & 0.1 & 11.6 & 10.8 & 75.5\\ \hdashline[2pt/2pt]
Microwave & ScrewNet & 0.79  & 0.81 & \textbf{0.13} & 0.52 & 1.19 & 0.54 & 0.01 & 0.01 & - & - & - & - & -\\ 
 & DUST-net & \textbf{0.41} & \textbf{0.42} & 0.22 & \textbf{0.43} & \textbf{0.46} & \textbf{0.40} & \textbf{0.00} & \textbf{0.00} & 0.7 & 0.6 & 19.7 & 14.3 & 39.9 \\ \hdashline[2pt/2pt]
Drawer & ScrewNet & 0.69 & \textbf{0.24} & 0.49 & \textbf{0.24} & \textbf{0.72} & 0.97 & 0.08 & \textbf{0.08} & - & - & - & - & - \\
& DUST-net & \textbf{0.42} & 0.50 & \textbf{0.32} & 0.74 & 0.75 & \textbf{0.56} &\textbf{ 0.07} & \textbf{0.08} & 0.2 & 0.1 & 12.3 & 31.6 & 55.2
\end{tabular}%
}
\vspace{2pt}
\caption{Mean error values on the MAAD and Screw Loss metric for estimation of articulation model parameters for real-world objects when network was trained solely using simulated data. ScrewNet predictions are reported in the camera frame. Angular values \{$\lhat, \hat{\m}, \theta_i, \theta_{\lhat}$\} and distances \{$\norm{\m}, D, d_i, d_\lhat$\} are reported in radian and meter, respectively. Numerical values are reported for the uncertainty parameters \{$\lambda_i, \beta_j$\}. Symbol $-$ represents value not reported.}
\label{tab:real-world}

\end{table}

\end{document}